%% file: main.tex
\documentclass[runningheads]{llncs}
\usepackage{graphicx}
\usepackage{tikz}
\usepackage{comment}
\usepackage{amsmath,amssymb} % define this before the line numbering.
\usepackage{color,wrapfig}
\usepackage{orcidlink}

\usepackage[capitalize]{cleveref}

\usepackage[accsupp]{axessibility}  % Improves PDF readability for those with disabilities.

\usepackage{graphicx}
\usepackage{amsmath}
\usepackage{amssymb}
\usepackage{booktabs}
\usepackage[labelfont=bf,list=true,font=small]{subcaption}
\usepackage[margin=0pt,labelfont=bf,font=small]{caption}
\usepackage{algorithm,algpseudocode} 
\usepackage{amsmath}
\usepackage{amssymb,xcolor}
\usepackage{gensymb,enumitem}

\begin{document}
% \linenumbers
\pagestyle{headings}
\mainmatter
\def\ECCVSubNumber{5778}  % Insert your submission number here

\title{Incremental Task Learning with \\ Incremental Rank Updates} % Replace with your title

\begin{comment}
\titlerunning{ECCV-22 submission ID \ECCVSubNumber} 
\authorrunning{ECCV-22 submission ID \ECCVSubNumber} 
%
\author{Anonymous ECCV submission}
\institute{Paper ID \ECCVSubNumber}
\end{comment}
%******************

% \begin{comment}
\titlerunning{Incremental Task Learning with Incremental Rank Updates}
\author{Rakib Hyder\inst{1}\orcidlink{0000-0003-4191-301X} \and Ken Shao\inst{1}\orcidlink{0000-0001-8249-1111} \and Boyu Hou\inst{1} \and
Panos Markopoulos\inst{2}\orcidlink{0000-0001-9686-779X} \and \\
Ashley Prater-Bennette\inst{3}\orcidlink{0000-0002-4272-423X} \and M. Salman Asif\inst{1}\orcidlink{0000-0001-5993-3903}
% \thanks{This paper is accepted in ECCV 2022.}
} 
% \email{\texttt{sasif@ece.ucr.edu}}
%
\authorrunning{R. Hyder et al.}
\institute{University of California Riverside \and Rochester Institute of Technology \and Air Force Research Laboratory}

%******************
\maketitle
\begin{abstract}

Incremental Task learning (ITL) is a category of continual learning that seeks to train a single network for multiple tasks (one after another), where training data for each task is only available during the training of that task. Neural networks tend to forget older tasks when they are trained for the newer tasks; this property is often known as catastrophic forgetting. To address this issue, ITL methods use episodic memory, parameter regularization, masking and pruning, or extensible network structures. In this paper, we propose a new incremental task learning framework based on low-rank factorization. In particular, we represent the network weights for each layer as a linear combination of several rank-1 matrices. To update the network for a new task, we learn a rank-1 (or low-rank) matrix and add that to the weights of every layer. We also introduce an additional selector vector that assigns different weights to the low-rank matrices learned for the previous tasks. We show that our approach performs better than the current state-of-the-art methods in terms of accuracy and forgetting. Our method also offers better memory efficiency compared to episodic memory- and mask-based approaches. Our code will be available at \url{https://github.com/CSIPlab/task-increment-rank-update.git}
\end{abstract}

\input{intro}

\input{background}

\input{technical}

\input{experiments}

\section{Conclusion} 
We proposed a new incremental task learning method in which we update the network weights using low rank increments as we learn new tasks. Network layers are represented as a linear combination of low-rank factors. To update the network for a new task, we freeze the factors learned for previous tasks, add a new low-rank (or rank-1) factor, and combine that with the previous factors using a learned combination. The proposed method offered considerable improvement in performance compared to the state-of-the-art methods for ITL in image classification tasks. In addition, the proposed low-rank ITL circumvents the use of memory buffer or large memory overhead while achieving  zero forgetting.

The need for task ID knowledge is a general limitation of our and other ITL methods. Such methods can be useful for incremental multitask learning where task ID is available during inference but training data is only available in a short window. Extending this method to class incremental learning (which does not require task ID) is an important problem for future work.

\noindent \textbf{Acknowledgments.} This material is based upon work supported in part by by Air Force Office of Scientific Research (AFOSR) awards FA9550-21-1-0330, FA9550-20-1-0039, Office of Naval Research (ONR) award N00014-19-1-2264, and National Science Foundation (NSF) award CCF-2046293. Approved for Public Release by AFRL; 
Distribution Unlimited: Case Number AFRL-2021-4063

\noindent \textbf{Corresponding author:} M. Salman Asif (sasif@ucr.edu)

\bibliographystyle{splncs04}
\bibliography{continual}

\end{document}

%% file: intro.tex
\section{Introduction} 
Deep neural networks have been extremely successful for a variety of learning and representation tasks (e.g., image classification, object detection/segmentation, reinforcement learning, generative models). A typical network is trained to learn a function that maps input to the desired output. The input-output relation is assumed to be fixed and input-output data samples are drawn from a stationary distribution \cite{parisi2019continual}. 
If the input-output relations or data distributions change, the network can be retrained using a new set of input-output data samples. 
Since the storage, computing, and network capacity are limited, we may need to replace old data samples with new samples. 
Furthermore, privacy concerns may also force data samples to be available for a limited time \cite{delange2021continual,parisi2019continual}. 
In such a training process, a network often forgets the previously learned tasks; this effect is termed \textit{catastrophic forgetting} \cite{mccloskey1989catastrophic,ratcliff1990connectionist}. 

\begin{figure*}[t]%\label{Intro}
      \centering
      \includegraphics[width=0.95\linewidth]{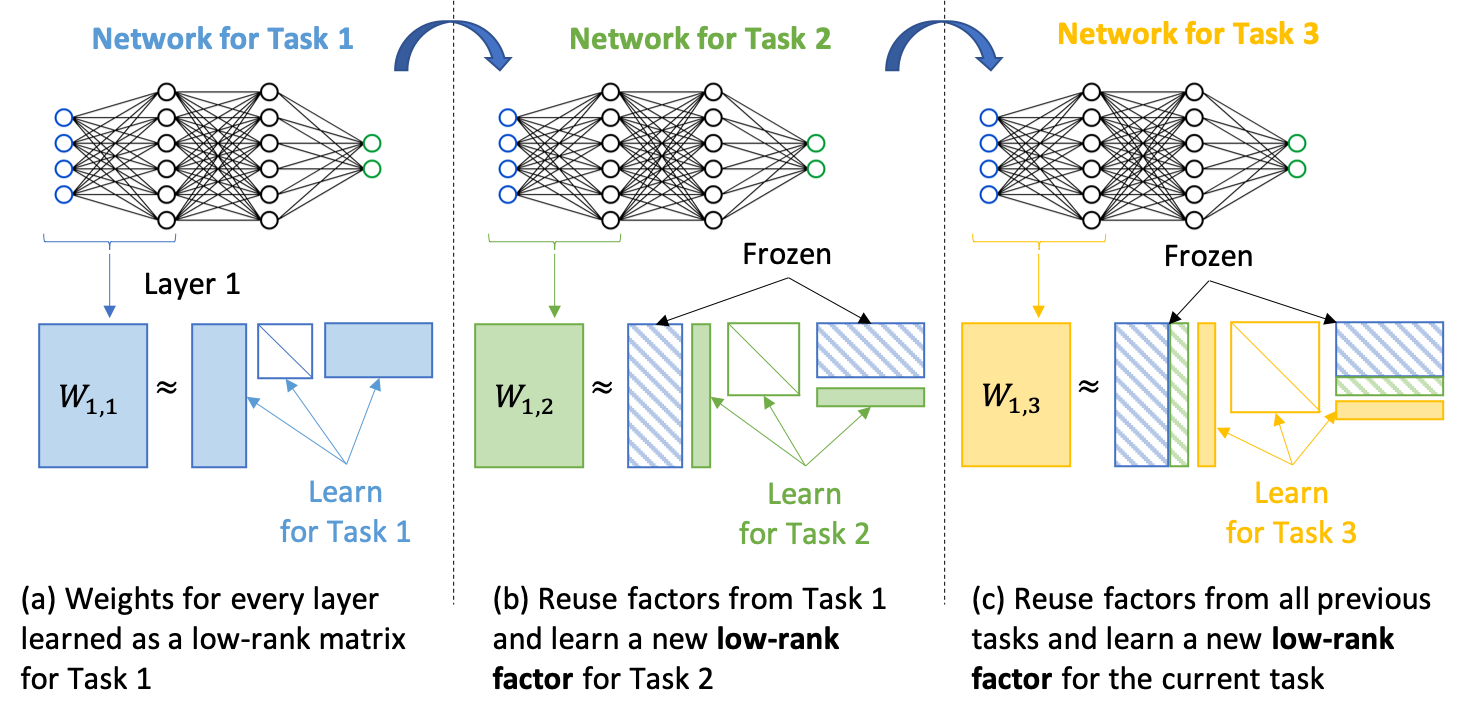}
      \caption{An overview of our proposed method for continual learning via low-rank network updates. We first represent (and learn) the weight matrix (or tensor) for each layer as a product of low-rank matrices. To train a network for new tasks without forgetting the earlier tasks, we reuse the factors from the earlier tasks and add a new set of factors for the new task. Our experiments suggest that a rank-1 update is often sufficient for successful continual learning.}
      \label{fig:intro}
      
\end{figure*}

% What is continual learning ? 
%Brief background on CL ...\\
%What has been done so far, what is the gap? \\
% 
Incremental task learning is a subcategory of continual learning or lifelong learning approaches aim to address the problem of catastrophic forgetting by adapting the network or training process to learn new tasks without forgetting the previously learned ones \cite{nguyen2017variational,li2017learning,aljundi2017expert,aljundi2018memory,aljundi2019gradient,chaudhry2018riemannian,riemer2018learning,rolnick2019experience,farajtabar2020orthogonal}. In this paper, we focus on task-incremental continual learning in which data for every task are provided in a sequential manner to train/update the network \cite{chaudhry2018efficient}. It has been a popular continual learning setup even in the very recent literature \cite{chaudhry2020continual,hurtado2021optimizing,deng2021flattening,veniat2020efficient,saha2020gradient,yin2021mitigating}. ITL finds its application in setups where task id is available during inference; for instance, tasks performed under different weather/light/background conditions and we know the changes, or tasks learned on different data/classes where we know the task id.

%
% Let us denote the network for task $T$ as $f_T($
%
Let us denote the network function that maps input $x$ to output for task $t$ as 
$f(x;\mathcal{W}_t)$, 
where $\mathcal{W}_t$ denotes the network weights for task $t$. We seek to update the $\mathcal{W}_t$ for all $t$ as we sequentially receive dataset for one task at a time. 
Suppose the training dataset for task $t$ is given as $(\mathcal{X}_t, \mathcal{Y}_t)$ drawn from a distribution  $\mathcal{P}_t$, where $\mathcal{X}_t$ denotes the set of input samples and $\mathcal{Y}_t$ denotes the corresponding ground-truth outputs. Our goal is to update network weights from the previous task ($\mathcal{W}_{t-1}$) to $\mathcal{W}_t$ such that 
\begin{equation}
    y \approx f(x; \mathcal{W}_t), \quad \text{for all } (x,y) \sim \mathcal{P}_t.
\end{equation}
ITL setup above assumes that the task identity of test samples is known at the test time and the corresponding network weights are used for inference. 
%This is in contrast to a more challenging case where we may need to identify the task along with the label \cite{wortsman2020supermasks}. 
%
Dynamic architecture approaches have the potential to achieve \textit{zero forgetting}, using $\mathcal{W}_t$ for testing data for task $t$; however, this also requires storing the $\mathcal{W}_t$ for all the tasks. One of the main contributions of this paper is to represent, learn, and update the $\mathcal{W}_t$ using low-rank factors such that they can be stored and applied with minimal memory and computation overhead.

%  Quick overview 
% - Low memory + computational overhead because we use a small number of parameters 
% - zero forgetting ? 
% TLDR of this paper 
We propose a new method for ITL that updates network weights using rank-1 (or low-rank) increments for every new task. Figure~\ref{fig:intro} provides an illustration of our proposed method. 
We represent the network weights for each layer as a linear combination of several low-rank factors (which can be represented as a product of two low-rank matrices and a diagonal matrix). 
To update the network for task $t$ without forgetting the earlier tasks, we freeze the low-rank factors learned from the previous tasks, add a new trainable rank-1 (or low-rank) factor for every layer, and combine that with the older factors using learnable \textit{selector weights} (shown as a diagonal matrix). We use a multi-head configuration that has an independent output layer for each task. As we are learning separate diagonal matrices for every task, we can achieve zero forgetting during inference.
We present an extensive set of experiments to demonstrate the performance of our proposed method for different benchmark datasets. We observe that our proposed method outperforms the current state-of-the-art methods in terms of accuracy with small memory overhead.

% Main contributions
The main contributions of this paper are as follows. 
\begin{enumerate}% [leftmargin=*,noitemsep,topsep=0pt]
    \item \textbf{Represent layers as low-rank matrices:} We represent and learn network weights for each layer as a low-rank structure. We show that low-rank structure is sufficient to represent all the tasks in continual learning setup.
    \item \textbf{Reuse old factors for better performance with a small memory overhead:} We limit the number of parameters required for network update by reusing the factors learned from previous tasks. We demonstrate that a rank-1 increment suffices to outperform the existing techniques.
    \item \textbf{Zero forgetting without replay buffer:} Our method has zero forgetting that is achieved using incremental rank update or network weights. In contrast, most of the existing continual learning techniques require replay buffer or large memory overhead to achieve zero forgetting. 
\end{enumerate}

% Discuss limitations 
% The main limitations of our approach are as follows. 
% \begin{enumerate}
%     \item 
% \end{enumerate}
\noindent \textbf{Limitations.} Our method shares same inherent limitation of ITL (i.e. the requirement of task-id during inference). In addition, since we use all the previously learned factors for inference, the later tasks require more memory and computation for inference. Nevertheless, we show that using low-rank structure, our total memory requirement is significantly lower than a single network. Furthermore, as we learn separate diagonal matrices for each task, we can maintain high performance even if the network reaches full rank with a large number of tasks.

%% file: background.tex
\section{Background and Related Work}
Incremental task learning (ITL) \cite{delange2021continual,silver2002task} aims to train a single model on a sequence of different tasks and perform well on all the trained tasks once the training is finished. While training on new tasks, the old data from previous tasks will not be provided to the model. This scenario mimics the human learning process where they have the ability to acquire new knowledge and skills throughout their lifespan. However, this setting is still challenging to neural network models as a common phenomenon called "catastrophic forgetting \cite{mccloskey1989catastrophic}" is observed during this learning process. Catastrophic forgetting occurs when the data from the new tasks interfere with the data seen in the previous tasks and thus deteriorating model performance on preceding tasks.
To overcome this issue, different approaches have been proposed so far which can be divided into three main categories: regularization-based approaches, memory and replay-based approaches, and dynamic network architecture-based approaches. Some of these approaches are especially designed for ITL whereas others are designed for more general continual learning setup.

\textbf{Regularization-based approaches} \cite{kirkpatrick2017overcoming,nguyen2017variational,li2017learning} update the whole model in each task but a regularization term $\ell_{reg}$ is added to the total loss $\mathcal{L} = \ell_{current} + \lambda \ell_{reg}$ to penalize changes in the parameters important to preceding tasks thus preserving the performance on previous learned tasks. For example, Elastic Weight Consolidation (EWC) \cite{kirkpatrick2017overcoming} estimates the importance of parameters using Fisher Information matrix; Variational Continual Learning (VCL) \cite{nguyen2017variational} approximates the posterior distribution of the parameters using variational inference; Learning without Forgetting (LwF) \cite{li2017learning} regularizes the current loss with soft targets taken from previous tasks using knowledge distillation \cite{hinton2015distilling}. GCL \cite{buzzega2020dark} mixes rehearsal with knowledge distillation and regularization to mitigate catastrophic forgetting. A number of recently proposed methods force weight updates to belong to the null space of the feature covariance  \cite{wang2021training,tang2021layerwise}. 

\textbf{Memory-based approaches} \cite{rebuffi2017icarl,riemer2018learning,chaudhry2018efficient,chaudhry2019tiny,tang2021layerwise} usually use memory and replay/rehearsal mechanism to recall a small episodic memory of previous tasks while training new tasks thus reduce the loss in the previous tasks. For example, iCaRL \cite{rebuffi2017icarl}  is the first replay method, which learns in a class-incremental way by selecting and storing exemplars closest to the feature mean of each class; Meta-Experience Replay (MER) \cite{riemer2018learning} combines experience replay with optimization-based meta-learning to optimize the symmetric trade-off between transfer and interference  by enforcing gradient alignment across examples; AGEM \cite{chaudhry2018efficient} projects the gradient on the current minibatch by using an external episodic memory of patterns from previous experiences as an optimization constraint; ER-Ring \cite{chaudhry2019tiny} jointly trains new task data with that of the previous tasks.

\textbf{Dynamic network architectures} \cite{rusu2016progressive,mallya2018packnet,wortsman2020supermasks,wen2020batchensemble,serra2018overcoming,chaudhry2020continual,yoon2018lifelong} try to add new neurons to the model at additional new tasks, thus the performances on previous tasks are preserved by freezing the old parameters and only updating the newly added parameters. For example, Progressive neural networks (PNNs) \cite{rusu2016progressive} leverage prior knowledge via lateral connections to previously learned features;  PackNet \cite{mallya2018packnet} iteratively assigns parameter subsets to consecutive tasks by constituting binary masks. SupSup \cite{wortsman2020supermasks} also finds masks in order to assign different subsets of the weights for different tasks. BatchEnsemble \cite{wen2020batchensemble} learns on separate rank-1 scaling matrices for each task which are then used to scale weights of the shared network. HAT \cite{serra2018overcoming} incorporates task-specific embeddings for attention masking. \cite{oswald2020continual} also proposes  task-conditioned hypernetworks for continual learning.
\cite{masse2018alleviating} proposes nonoverlapping sets of units being active for each task. Piggyback \cite{mallya2018piggyback} learns binary masks on an existing network to provide good performance on new tasks. \cite{abati2017conditional} proposes task specific convolutional filter selection for continual learning. 
The mask-based methods listed above provide excellent results for continual learning, but they require a significantly large number of parameters to represent the masks for each task. 
A factorization-based approach was proposed in \cite{mehta2021continual} that performs automatic rank selection per task for variational inference using Indian Buffet process. The method requires significantly large rank increments per task to achieve high accuracy; in contrast, our method uses a learning-based approach to find rank-1 increments and reuse old factors with the learned selector weights. ORTHOG-SUBSPACE \cite{chaudhry2020continual} learns tasks in different (low-rank) vector sub-spaces that are kept orthogonal to each other in order to minimize  interference. 

Our proposed method falls under the category of dynamic network architecture approaches. Note that we can represent a low-rank weight matrix using two smaller fully-connected layers and increasing the rank of the weight matrix is equivalent to adding new nodes in the two smaller  fully-connected layers.

%% file: technical.tex
\section{Incremental Task Learning via Rank Increment}

We focus on the incremental task learning setup in which we seek to train a network for $T$ tasks. The main difference between incremental task learning and regular learning is that the training data for every task is only available while training the network for that task. The main challenge in incremental task learning is to not forget the previous tasks as we learn new tasks. 
Learning each task entails training weights for the network to learn the task-specific input-output relationship using the task-specific training data.

We seek to develop an ITL framework in which we represent the weights of any layer using a small number of low-rank factors. We initialize the network with a base architecture in which weights for each layer can be represented using a low-rank matrix. We then add new low-rank factors to each layer as we learn new tasks.

Let us assume the network has $K$ layers and the weights for the $k$th layer and task $t$ can be represented as $W_{k,t}$. Let us further assume that the weights for the $k$th layer and task $t= 1$ can be represented as a low-rank matrix
\begin{equation}
    W_{k,1} = U_{k,1}S_{k,1,1}V_{k,1}^\top, 
\end{equation}
where $U_{k,1},V_{k,1}$ represent two low-rank matrices and $S_{k,1,1}$ represents a diagonal matrix. To learn the network for task 1, we learn $U_{k,1},V_{k,1},S_{k,1,1}$ for all $k$. For task 2, we represent the weights for $k$th layer as 
$$W_{k,2} = U_{k,1}S_{k,1,2}V_{k,1}^\top + U_{k,2}S_{k,2,2}V_{k,2}^\top.$$ 
$U_{k,1},V_{k,1}$ represent the two low-rank matrices learned for task 1 and frozen afterwards.  $U_{k,2},V_{k,2}$ represent two low-rank matrices that are added to update the weights, and these will be learned for task 2.  $S_{k,1,2}, S_{k,2,2}$ represent the diagonal matrices, which will be learned for task 2. We learn $S_{k,1,2}$, which is a diagonal matrix that assigns weights to factors corresponding to task 1, to include/exclude or favor/suppress  frozen factors from previous tasks for the new tasks. 
We can represent the weights for the $k$th layer and task $t$ as 
\begin{align}
    W_{layer,task} & = W_{k,t} = \sum_{i\le t} U_{k,i} S_{k,i,t} V_{k,i}^\top \nonumber\\&= \sum_{i<t} {\color{red} \underbrace{U_{k,i}}_\text{frozen}} S_{k,i,t} {\color{red}\underbrace{V_{k,i}^\top}_{\text{frozen}} }  + U_{k,t} S_{k,t,t} V_{k,t}^\top, \label{eq:weight} 
\end{align} 
where $U_{k,i},V_{k,i}$ are frozen for all $i<t$ and $U_{k,t},V_{k,t}$ and all the $S_{k,i,t}$ are learned for task $t$. The entire network for task $t$ can be represented as $\mathcal{W}_t = \{U_{k,i},S_{k,i,t},V_{k,i}\}_{i\le t}$. 
To update the trainable network parameters for task $t$, we solve the following optimization problem:
\begin{align}
    % \scriptstyle
    \underset{U_{k,t},S_{k,i,t},V_{k,t}}{\min} &\sum_{(x,y)\in (\mathcal{X}_t,\mathcal{Y}_t)} \text{loss}(f(x;\mathcal{W}_t[U_{k,t},S_{k,i,t},V_{k,t}]), y)\notag \\
     &\text{for all }  k\leq K \text{ and } i\leq t,
    \label{eq:parameter_opt}
\end{align}
where we use $\text{loss}(\cdot,\cdot)$ to denote the loss function and $\mathcal{W}_t[U_{k,t},S_{k,i,t},V_{k,t}]$ to indicate the trainable parameters in $\mathcal{W}_t$, while the rest are frozen. We sometimes call $S_{k,i,t}$ a \textit{selector weight matrix/vector} to indicate that its diagonal entries determine the contribution of each factor toward each task/layer weights. 

Our proposed ITL algorithm works as follows. We train the low-rank factors for the given task using the respective training samples. Then we freeze the factors corresponding to the older tasks and only update the new factors and the diagonal matrices. In this manner, the total number of parameters we add in our model is linearly proportional to the rank of the new factors. To keep the network complexity small, we seek to achieve good accuracy using small rank for each task update and layer. We summarize our approach in Algorithms~\ref{algo:train} and~\ref{algo:test}.

Note that we do not need to create the weight matrix $W_{k,t}$ for any layer explicitly since we can compute all the steps in forward and backward propagation efficiently using the factorized form of each layer. 
The size of each layer is determined by the choice of the network architecture. The rank of each layer for every task is a hyper-parameter that we can select according to the tasks at hand. To keep the memory overhead small, we need to use small values for rank increment. Let us denote the rank for $U_{k,t} $ as $r_{k,t}$, which represents the increment rank for $k$th layer and task $t$. At the time of test, we can use an appropriate number of factors depending on the task. For instance, if we want to predict output for task 1 then we use first $r_{k,1}$ factors and for task 2 we use $r_{k,1}+r_{k,2}$ factors. We can add new factors in an incremental manner as we add new tasks in the ITL setup. In the extreme case of rank-1 increments, $r_{k,t} = 1$. In our experiments, we observed that rank-1 updates compete or exceed the performance of existing ITL methods (see Table~\ref{table:compare}) and the performance of our method improves further as we increase the rank (see Table~\ref{table:rank}). Any increase in the rank comes at the expense of an increased memory overhead.

\begin{algorithm}[t]
   \caption{ITL with rank-1 increments (Training)}
   \label{algo:train}
\begin{algorithmic}
    
    \State {\bfseries Input:} Data ($\mathcal{X}_1$ and $\mathcal{Y}_1$) for the $1^{st}$ task.
    \State Set initial rank, $r_1$.
    \State Initialize weight factors $U_{k,1},V_{k,1}$ at random and $S_{k,1,1}$ as an identity marix. 
    % \State Create weight matrices. \Comment{ using \eqref{eq:weight}}
    % \State Calculate forward pass.  % \Comment{using \eqref{eq:input_output}}
    \State Learn $U_{k,1},V_{k,1}$ and $S_{k,1,1}$. \Comment{Optimization in  \eqref{eq:parameter_opt}}
    % \State \texttt{\color{blue}\% Comments???}
    
    \For{$t = 2, 3, ..., T$}
        \State {\bfseries Input:}  Training data ($\mathcal{X}_t$ and $\mathcal{Y}_t$) for $t^{th}$ task.
        \State Initialize low-rank update factors $U_{k,t},V_{k,t}$.
        \State Freeze the previous factors $\{U_{k,i},V_{k,i}\}_{i<t}$. 
        \State Initialize the diagonal entries of $\{S_{k,i,t}\}$ as 1 
        \State \;\;\;\;for $i=t$ and 0 for $i<t$. 
        % \State Create weight matrices. \Comment{ Using \eqref{eq:weight}}
        % \State Calculate forward pass.  \Comment{Using \eqref{eq:input_output}}
        \State Learn $U_{k,t},V_{k,t}$ and $S_{k,i,t}$ 
        \State   \;\;\;\;for $i<t$. \Comment{Optimization in \eqref{eq:parameter_opt}}
        
   \EndFor
\end{algorithmic}

\end{algorithm}

\begin{algorithm}[t]
   \caption{ITL with rank-1 increments (Inference)}
   \label{algo:test}
\begin{algorithmic}
    
    \State {\bfseries Input:} Test data $x$ with task identity $t$. \\
    {\bfseries Retrieve trained weights:}  $\mathcal{W}_t = \{U_{k,i},V_{k,i},S_{k,i,t}\}$ for all $k$ and $i\le t$. 
    % \State Create weight matrices. \Comment{ Using Eqn.~\ref{eq:weight}} 
    \State {\bfseries Output:} Calculate the network output as $f(x, \mathcal{W}_t)$. 
    % inference using the created weight matrices.  \Comment{Using Eqn~\ref{eq:input_output}}
\end{algorithmic}
\end{algorithm}

%% file: experiments.tex
\section{Experiments and Results}
% \Salnote{start filling this part}
We used different classification tasks on well known continual learning benchmarks to show the significance of our proposed approach. 

\subsection{Datasets and Task Description}
Experiments are conducted on four datasets: Split CIFAR100,  Permuted MNIST, Rotated MNIST, and Split MiniImageNet.\\
 \textbf{P-MNIST} creates new tasks by applying a certain random permutation on the pixels of all images in the original dataset. In our experiment, we generate 20 different tasks, each of which corresponds to a certain but different permutation.\\
 \textbf{R-MNIST} is similar to Permuted MNIST, but instead of applying a certain random permutation on the pixels, it applies a certain random rotation to the images in the same tasks. We create 20 different tasks, each corresponds to a certain but different version of rotation from [0, 180] degree interval.\\
 \textbf{S-CIFAR100} splits the original CIFAR-100 dataset into 20 disjoint sets, each of which, containing 5 classes, is considered as a separate task. The 5 classes in each task is randomly chosen without replacement from the total 100 classes.\\
\textbf{S-miniImageNet} splits a subset of Imagenet dataset into 20 disjoint sets, each of which, containing 5 classes, is considered as a separate task. The 5 classes in each task is randomly chosen without replacement from the total 100 classes.

\subsection{Training Details}

\noindent \textbf{Network.} In the first set of experiments, we used a three layer (fully-connected) multilayer perceptron (MLP) with 256-node hidden layers, similar to the network in \cite{chaudhry2020continual}. We flattened multi-dimensional input image to a 1D vector input. We used ReLU activation for all the layers except the last one. We used Softmax for the muticlass classification tasks. 
We used the same network for all the tasks with necessary modifications for input and output sizes. Our approach can be used in convolutional networks as well. We report the results using ResNet18 with our approach on S-CIFAR100 and S-miniImageNet dataset in Table~\ref{table:resnet}. 

% \noindent \textbf{Factorization.} We used the matrix factorization defined in \eqref{eq:weight} in all our experiments. We used the same initial rank or rank increment for all the layers except the last layer. 

% We reshape the convolutional filters for each layers of ResNet as 2D matrices with number of filters in one dimension and collapsing height, width, and the input channel in another. 

% \textbf{Incremental rank update:} For the first task, we allowed much higher rank without sacrificing any performance. For the subsequent tasks, we started with a much lower rank tensor cores and increased the rank depending on the performance. We added the choice of ranks for each task in supplementary material.

\noindent \textbf{Factorization and rank selection.} We used the matrix factorization defined in \eqref{eq:weight} in all our experiments. We empirically selected the rank for the first task,$r_{k,1}$ as 11 based on the experiments on a sample Rotated MNIST task and kept the same value for all the experiments. We then performed rank-1 increment ($r_{k,t}$) for each additional task. We would like to point that AGEM and Orthog Subspace use first 3 tasks for hyperparameter tuning. We did not tune our hyperparameters on the test data, rather we choose the parameters which provides better convergence during training. We increment the weight matrices by rank-1 per task; therefore, learning rate and the number of epochs are the only  hyperparameters in our experiments.

\noindent \textbf{Optimization.} We used orthogonal initialization for the low-rank factors, as described in \cite{saxe2013exact}. We used all one initialization for the additional factors of the selector matrices $S_{k,t,t}$. 
% We used mean squared error (MSE) as the loss function in our experiments. 
We used Adam optimization to update the factors. We used the batch size of 128 for each task.

\noindent \textbf{Performance metrics.} We use \emph{accuracy} and \emph{forgetting} per task, which are two commonly used metrics in the continual learning literature~\cite{chaudhry2018riemannian,chaudhry2020continual}, to evaluate the performance of the described methods.
Let $a_{t,j}$ be the test accuracy of task $j<t$ after the model has finished learning task $t\in\{1,...,T\}$ in a incremental manner. The average accuracy $A_{t}$ after the model has learned task $t$ is defined as
$ A_{t} = \frac{1}{t}\sum_{j = 1}^{t} a_{t,j}.$
% \begin{equation}
%     A_{t} = \frac{1}{t}\sum_{j = 1}^{t} a_{t,j}. \label{eq:acc} 
% \end{equation}
%
%
On the other hand, \emph{forgetting} is the decrease in the accuracy of a task after its training, and after one or several tasks are learned incrementally. We define the average forgetting $F_{t}$ as
$    F_{t} = \frac{1}{t-1}\sum_{j = 1}^{t-1} (a_{j,j} - a_{t,j}). $
%
% \begin{equation}
%     F_{t} = \frac{1}{t-1}\sum_{j = 1}^{t-1} (a_{j,j} - a_{t,j}). \label{eq:forget} 
% \end{equation}
%

In Figure~\ref{fig:task}, we show the evolution of average accuracy $A_{t}$ as $t$ increases. We also show the evolution of task-wise accuracy $a_{t,j}$ in Figure~\ref{fig:transition}, where $(t,j)$ pixel intensity reflects $a_{t,j}$. We report the average accuracy $A_{T}$, the average accuracy after the model has learnt every tasks incrementally, in Table~\ref{table:compare}.
We report the forgetting $F_{T}$ after the model has learnt all the tasks incrementally in Table~\ref{table:forgetting}. Note that our method performs incremental task learning without forgetting. 
% and the task-wise transfer of knowledge is hidden in the evolution of the selector matrices.

\subsection{Comparing Techniques}

We compare our method against different state-of-the-art ITL methods. \textbf{EWC} \cite{kirkpatrick2017overcoming} is a regularization-based method that uses the Fisher Information matrix to estimate posterior of previous tasks to preserve important parameters. \textbf{ICARL}~\cite{rebuffi2017icarl} is a memory-based method that uses exemplars and knowledge distillation \cite{hinton2015distilling}
to retain previous knowledge. \textbf{AGEM} \cite{chaudhry2018efficient} is a memory-based method built upon \cite{lopez2017gradient} that uses episodic memory to solve an constrained optimization problem.  \textbf{ER-Ring} \cite{chaudhry2019tiny} is another memory-based method that jointly trains on new task data with that of the
previous tasks. \textbf{Orth. sub.} \cite{chaudhry2020continual} learn tasks in different (low-rank) vector subspaces that are kept orthogonal to each other in order to minimize interference. Other than the above mentioned approaches, we compared with masked based approaches which, like our approach, also fall under dynamic architecture category.  \textbf{HAT}~ \cite{serra2018overcoming} that incorporates task-specific embeddings for attention masking. \textbf{PackNet}~ \cite{mallya2018packnet} that iteratively assigns subsets of a single binary mask to each task. 
%\textbf{SupSup}~ \cite{wortsman2020supermasks} that learns one mask for each task. 
The mask-based approaches utilize the redundancy of the network parameters to represent different tasks with different masked versions of the same network weights.
We also present comparisons with some recent methods: \textbf{IBP-WF} \cite{mehta2021continual} and \textbf{Adam-NSCL} \cite{wang2021training}, in terms of average accuracy for one experiment on two datasets. 

In addition, we report results for two \textit{non-continual} baseline methods: \textbf{Parallel learning} and \textbf{Multitask learning}. \textbf{Parallel learning} trains independent (smaller) low-rank networks of same size for each task. We report results for three such networks. \textbf{Parallel 2} uses rank-2 layers, \textbf{Parallel 4} uses rank-4 layers, and \textbf{Parallel full} uses a full-rank MLP. Parallel 2 requires approximately the same number of parameters as the rank-1 ITL network that we use in our experiments; Parallel 4 provides higher network capacity, while requiring fewer parameters than the full-rank network. We can treat the performance of the \textbf{Parallel full} approach as the upper limit that we can achieve using ITL methods. Finally, \textbf{Multitask learning} has been used as a baseline in \cite{chaudhry2020continual,chaudhry2018efficient}. In multitask learning, we have access to all data to optimize a single network.

\begin{table*}[t]
\caption{Average test accuracy of ITL for P-MNIST, R-MNIST, S-CIFAR100, and S-miniImageNet with three layer MLP. Standard deviation of test accuracy over five runs is shown in parenthesis for some of the experiments. \\[1ex]
$^*$ Orthog subspace does not use replay buffer for MNIST variations.\\[-1.5 ex]}
\label{table:compare}
% title name of the table
\small
%\footnotesize
\centering % centering table
\begin{tabular}{l|c|cccc}
\hline
\hline 
Method                      & \begin{tabular}[c]{@{}c@{}}Replay\\ Buffer\end{tabular} & P-MNIST           & R-MNIST                    & S-CIFAR100                              & S-miniImageNet                           \\ 
\hline 
{EWC} \cite{kirkpatrick2017overcoming}            & No                                                       & 67.9 ($\pm$0.68)  & 44.5 ($\pm$1.09)  & 52.7 ($\pm$0.81)                         & 29.3 ($\pm$1.08)                          \\\hline
{ICARL}  \cite{rebuffi2017icarl}            & Yes                                                      & 85.4 ($\pm$0.01)  & 51.2($\pm$ 2.41)                 & 56.9($\pm$0.31)                          & 39.9($\pm$0.27)                               \\\hline
{AGEM}   \cite{chaudhry2018efficient}           & Yes                                                      & 73.9 ($\pm$0.52)  & 53.4 ($\pm$1.80)  & 51.3($\pm$1.54)                          & 31.3($\pm$0.89)                       \\\hline
{HAT} \cite{serra2018overcoming}         &  No     &\textbf{90.1}($\pm$1.60)&89.1($\pm$2.51)&64.8 ($\pm$0.32) &47.0 ($\pm$0.88)       \\\hline
{PackNet} \cite{mallya2018packnet}   &  No         &90.0($\pm$0.24)&88.4($\pm$0.37)&63.7($\pm$0.41) & 45.1($\pm$1.05)                  \\\hline
%\textbf{SupSup}           &  No  &96.6($\pm$0.0047)&96.57($\pm$0.045)&68.8($\pm$2.04) &64.2($\pm$0.88)                       \\\hline
%
{Orth sub} \cite{chaudhry2020continual}         & Yes$^*$                                                    & 86.6  ($\pm$0.79) & 80.2 ($\pm$0.41)  & 57.8 ($\pm$1.03)                         & 38.1 ($\pm$0.67)                       \\\hline
%
% \textbf{DER} \cite{buzzega2020dark} & Yes & -- & -- & 48.21 & 33.19 
%  \\\hline
%

%
{Adam-NSCL}\cite{wang2021training} & No & -- & -- & 64.26 & 47.32
 \\\hline
 {IBP-WF} \cite{mehta2021continual} & No & -- & -- & 53.22 & 40.52
 \\\hline
{Ours}               & No                                                       & 85.6 ($\pm$0.15)  & \textbf{91.1} ($\pm$0.12)  & \textbf{65.9} ($\pm$2.16)                         &  \textbf{54.7} ($\pm$2.87)                        \\ \hline \hline
{Parallel 2  (r=2) }   & -                                                        & 65.3              & 65.5                          & 62.8                                     & 55.4                                     \\ \hline
{Parallel 4 (r=4) }   & -                                                        & 86.3              & 87.4                          & 65.6                                     & 58.6                                     \\  \hline
{Parallel fullrank} & -                                                        & 95.9              & 97.3                         & 73.1                                     & 63.1                                     \\ \hline
{Multitask }  & -                                                        & 96.8              & 97.7                            & 16.4 & 4.21 \\\hline

% \hline % inserts single-line
% \multicolumn{2}{l}{\textsuperscript{R} Uses Resnet18 architecture} \\
\end{tabular}
\end{table*}

\subsection{Results with Three-Layer MLP}

\noindent \textbf{Classification performance and comparison. } We report classification results for P-MNIST, R-MNIST, S-CIFAR100, and S-miniImageNet tasks in Table~\ref{table:compare}. We also show the results for the comparing techniques. We observe that our method with rank-1 update perform better than all the comparing methods (EWC, ICARL, AGEM, HAT, PackNet, Orthog Subspace) on R-MNIST, S-CIFAR100 and S-miniImageNet tasks using significantly fewer number of parameters. Our method performs close to  Orthog Subspace on P-MNIST tasks.

\begin{table*}[t]
\caption{Average forgetting results corresponding to Table~\ref{table:compare} for different datasets using different approaches. We report the forgetting in percentage unit (\%). We also report the standard deviation over 5 experiments for some methods. }
%\\ [2.5ex]
\label{table:forgetting}
\centering % centering table
\small
%\footnotesize
\begin{tabular}{l |c c c c  c} % columns
\hline % inserting double-line
\hline
% \\ [-2 ex]
  & EWC & AGEM &\begin{tabular}[c]{c}Orthog\\subspace  \end{tabular}  & \begin{tabular}[c]{c}Adam-\\NSCL \end{tabular}& \begin{tabular}[c]{c}Parallel fullrank,\\HAT, PackNet \\ \textbf{Ours}, IBP-WF\end{tabular}\\
  \hline

P-MNIST & 25.8 ($\pm$0.70) & 19.6 ($\pm$0.64) & 4.49 ($\pm$0.93) & -  & 0\\
R-MNIST & 52.9 ($\pm$1.17)  & 44.2 ($\pm$1.85) & 14.7 ($\pm$0.39) & - & 0\\
S-CIFAR100 & 6.96 ($\pm$0.80)  & 21.5 ($\pm$2.89) & 6.30 ($\pm$0.38) & 8.5 & 0\\

S-miniImageNet & 17.3 ($\pm$1.81)  & 18.8 ($\pm$1.40) & 9.98 ($\pm$0.31) & 11.23 & 0\\
\hline
\end{tabular}

\end{table*}

\begin{figure*}[h]
    \centering
    \includegraphics[width=0.9\linewidth]{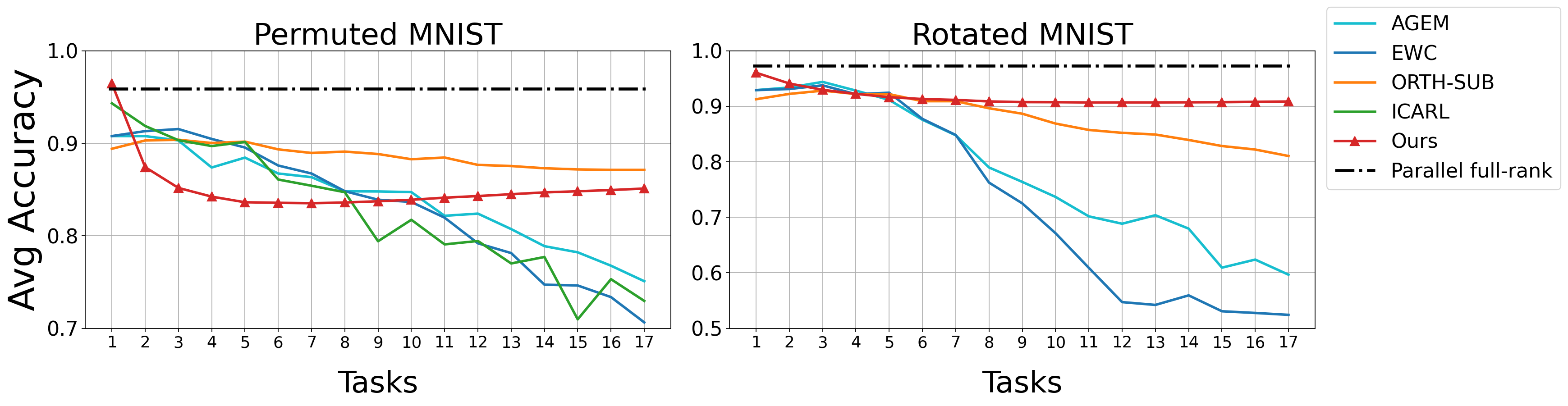}
    
    \includegraphics[width=0.9\linewidth]{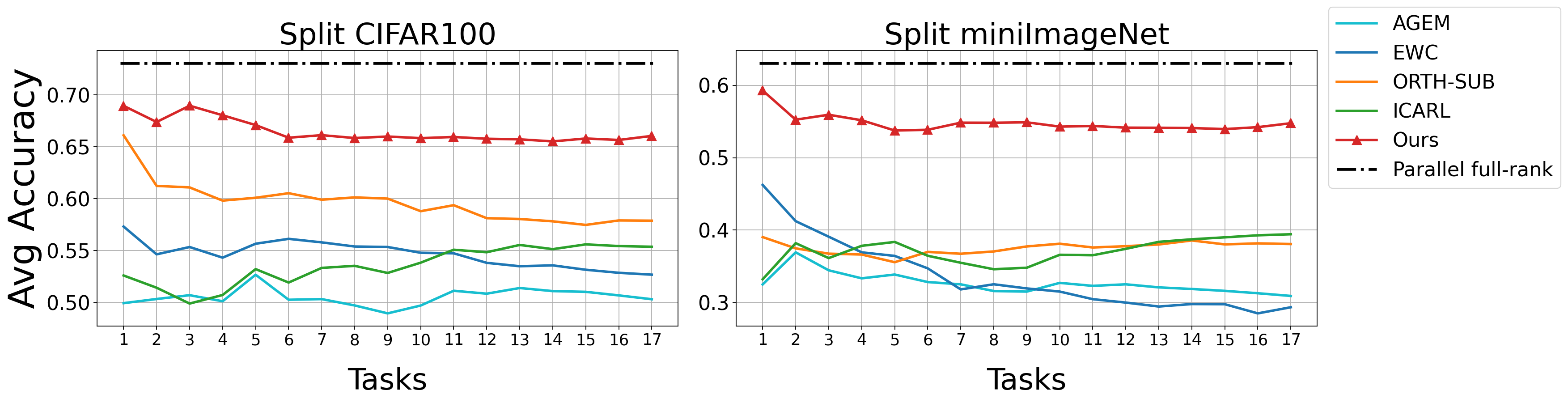}
    \caption{Average test accuracy for different datasets (Permuted MNIST, Rotated MNIST, Split CIFAR100, Split miniImageNet) along different tasks using different algorithms (AGEM,EWC, Orthog. Subspace, ICARL and our approach). We use three layer MLP here. Parallel full-rank results corresponds to the case when we train every task on separate full rank networks independently (serves as an upper limit for ITL methods). We showed the average of 20 tasks. }
    \label{fig:task}
\end{figure*}

\begin{figure}[t]
\centering
\includegraphics[width=0.85\linewidth]{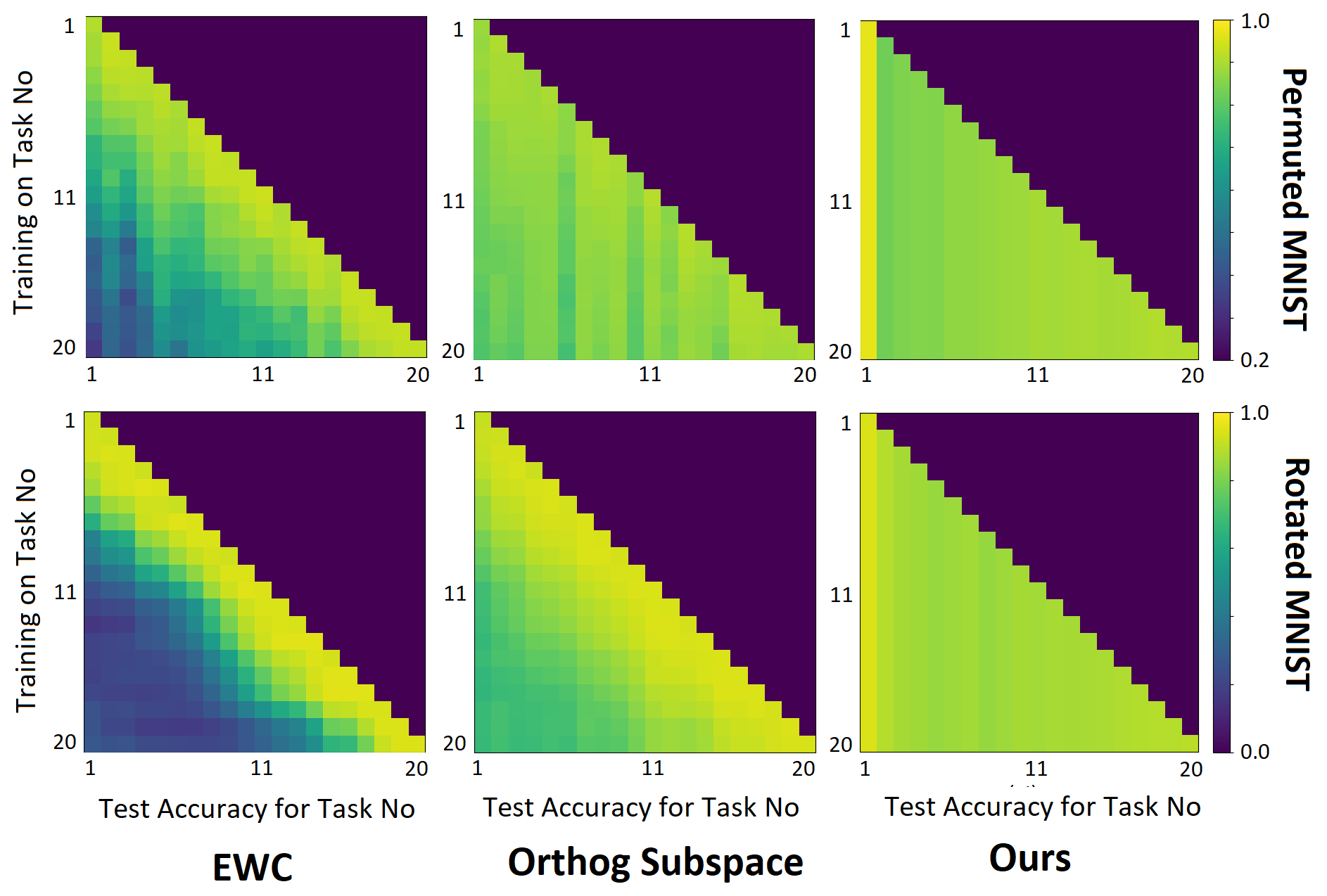}
\caption{ Evolution of task-wise test accuracy on P-MNIST (first row)  and R-MNIST  (second row)  datasets for EWC, Orthogonal Subspace, and Our approach. We can observe from the decrease in the test accuracy that EWC and Orthogonal Subspace forget the previous tasks as they learn new tasks. Our approach does not show any forgetting as we learn new tasks.}
\label{fig:transition}
\end{figure}

We also observe that the proposed rank-1 update outperforms non-continual Parallel 2 baseline that has similar number of parameters compared to our approach. We perform similar to Parallel 4 baseline that uses nearly twice the number of parameters as our approach. Parallel full acts as an upper limit with the network structure of our choice as it trains independent full rank networks for every task. Multitask learning is another non-continual baseline that uses all the data from all the tasks simultaneously.  Table~\ref{table:compare} suggests that our ITL method can learn complex tasks such as CIFAR100 and miniImageNet classification with a three layer MLP, whereas multitask learning (which is solving 100-class classification problem) fails with such a simple network. We also tested Resnet18 network, which has significantly more parameters than the network used in Table~\ref{table:compare}. The results for Resnet18 are presented in Table~\ref{table:resnet}. 

We present the task-wise test performance for some of the comparing approaches on P-MNIST, R-MNIST, S-CIFAR100 and S-miniImageNet datasets in Figure~\ref{fig:task}. We observe that as we train new tasks, task-wise performance drops for the comparing approaches, especially for P-MNIST and R-MNIST.

ICARL and AGEM require replay buffer (episodic memory) for each task. Although Orthog Subspace did not use replay buffer for MNIST experiments, it requires replay buffer in their algorithm and used it for S-CIFAR100 and S-miniImageNet experiments. EWC does not require any replay buffer, but it suffers from high forgetting as shown in Figure~\ref{fig:transition}. Our proposed approach does not require a replay buffer, and it outperforms other approaches in Table~\ref{table:compare}.

\noindent \textbf{Accuracy vs forgetting.} We report the average forgetting of different comparing approaches in Table~\ref{table:forgetting}. Our method, mask-based approaches (HAT and PackNet) and parallel baselines have zero forgetting, whereas all other comparing methods exhibit some level of forgetting.
To better demonstrate the forgetting, in Figure~\ref{fig:transition}, we show the accuracy for the tasks along the entire training procedure. $i^{th}$ row (top-bottom) of the diagram denotes the performance of $i$ tasks on the test sets when we train the $i^{th}$ task. As expected, we can observe that the training performance for the previously learned tasks usually drops with the gradual training of the subsequent tasks specially for the regularization based approach, EWC. However, our algorithm maintains the same performance for the past tasks as we do not change any previously learned factors. Even orthogonal subspace approach observes such forgetting over some tasks.

\noindent \textbf{Memory complexity.} Our method increments the rank of each layer for each task; therefore, we compare the total number of parameters in the incrementally trained network and the Parallel baselines. Note that if the number of parameters in two approaches is same, we can train one small network per task independently. We report total number of parameters and replay buffer size for different methods in Table~\ref{table:memory}. Since we used similar fully connected network structure for all the tasks, we report results for Split CIFAR100 experiments. Although we increase the rank for every task, the increment is small enough that even after 20 tasks our total parameter count remains smaller than all other methods. 

% \begin{small} 
% \centering 
% \begin{tabular}{l |c c c c c c c} % columns
% \hline % inserting double-line
% \hline
% \\ [-2 ex]
%   & \textbf{Ours}  & IBP-WF & Adam-NSCL & EWC &  AGEM & Ortho Sub & Parallel   fullrank
% \\ [0ex]
% \hline % inserts single-line
% % {{\#} parameters} & 0.93M (1.00) & \textbf{0.17M (0.21)}  & 1.76M (1.88) & 2.82M (3.03) & 19.7M (21.18) \\ \hline
% {{\#} params.} & \textbf{0.17M} & 0.23M & 0.88M &  0.93M & 1.76M  & 2.82M &  19.7M \\
% \hline 
% \end{tabular}
% \end{small}  
% %

\begin{table*}[t]
% \vspace{-4mm}
\caption{Number of parameters and buffer size in ITL methods with  3-layer MLP. 
%The numbers inside bracket is relative to the number of parameters of a single full-rank 3-layer MLP. 
% We use a replay buffer of 20 examples per class per task for AGEM and Orthog. subspace.
}
\label{table:memory}
\centering % centering tabls
\small
\footnotesize
\begin{tabular}{l |c c c c c c c} % columns
\hline % inserting double-line
\hline
\\ [-2 ex]
   & {Ours} & IBP-WF & EWC &  AGEM & Ortho Sub    & Adam-NSCL & Parallel   fullrank
\\ [0ex]
\hline % inserts single-line
% {{\#} parameters} & 0.93M (1.00) & \textbf{0.17M (0.21)}  & 1.76M (1.88) & 2.82M (3.03) & 19.7M (21.18) \\ \hline
{{\#} params.} &  \textbf{0.17M} & 0.23M & 0.93M & 1.76M  & 2.82M  & 0.88M & 19.7M \\ \hline
{{buffer size} }& 0 & 0 & 1.71M &  7.90M & 9.01M & 0  & 0\\
\hline 
\end{tabular}
% \\[-3.0ex]
\end{table*}

% $2\times$ smaller than IBP-WF, $14\times$ smaller than AGEM, and $140\times$ smaller than the non-continual parallel full rank baseline.
%
% We also report the memory overhead in Table~\ref{table:memory} for different approaches. To calculate memory overhead, we combine incremental network/task head size and the size of replay buffer (i.e. the episodic memory). Since parallel full-rank network is non-continual, the memory overhead is not reported. We observe that our memory overhead is also significantly smaller than that of comparing approaches.

% \begin{table*}[h]
% % \vspace{-4mm}
% \caption{Total number of parameters required by different approaches for incrementally learning all 20 tasks on Split-CIFAR100 using 3-layer MLP. 
% %The numbers inside bracket is relative to the number of parameters of a single full-rank 3-layer MLP. 
% We use a replay buffer of 20 examples per class per task for AGEM and Orthog. subspace.}
% \label{table:memory}
% \centering % centering tabls
% \small
% \footnotesize
% \begin{tabular}{l |c c c c c c c} % columns
% \hline % inserting double-line
% \hline
% \\ [-2 ex]
%   & \textbf{Ours}  & IBP-WF & Adam-NSCL & EWC &  AGEM & Ortho Sub & Parallel   fullrank
% \\ [0ex]
% \hline % inserts single-line
% % {{\#} parameters} & 0.93M (1.00) & \textbf{0.17M (0.21)}  & 1.76M (1.88) & 2.82M (3.03) & 19.7M (21.18) \\ \hline
% {{\#} params.} & \textbf{0.17M} & 0.23M & 0.88M &  0.93M & 1.76M  & 2.82M &  19.7M \\
% \hline 
% \end{tabular}
% % \\[-3.0ex]
% \end{table*}

% \begin{wraptable}{r}{7.5cm}

%
We also report the number of parameters used by mask-based zero forgetting algorithms (HAT and PackNet) to learn 20 different tasks on different datasets in Table~\ref{table:param_mask}. We can observe that our approach outperforms HAT and PackNet for R-MNIST, S-CIAR100 and S-miniImageNet with a significantly smaller number of parameters. Even though all the approaches use the same network, our approach uses rank-1 factors that require a significantly smaller number of parameters for incremental learning of tasks. Note that P-MNIST and R-MNIST experiments require the same number of parameters. 

\begin{table}[t] 
\caption{Number of parameters used by different zero-forgetting algorithms (HAT, PackNet, and Ours) using 3-layer MLP. }
\label{table:param_mask}
\centering % centering tabls
\small
%\footnotesize
\begin{tabular}{l |c c c c  } % columns
\hline % inserting double-line
\hline
% \\ [-2 ex]
Method  & P/R-MNIST  & S-CIFAR100 & S-miniImageNet\\
% \\ [0ex]
\hline % inserts single-line
%SupSup &5.64M&17.92M&115.2M\\ \hline
HAT  & 0.33M&0.89M&5.51M  \\ \hline
PackNet &0.26M&0.83M&5.50M\\ \hline
Ours &0.11M&0.17M&0.72M \\ 
 \hline
\end{tabular}
%\\[-3.0ex]
\end{table}

\noindent \textbf{Effect of rank.} In Table~\ref{table:rank}, we evaluate the effect of different rank selection for different MNIST datasets using our ITL approach. We tested the initial rank (rank for the first task) of 1, 6, and 11, keeping the rank increment to 1. We observed that the accuracy increase as the initial rank increases, and we achieve nearly 90\% accuracy with initial rank of 11. We also tested different values of rank increment per task and observe that the accuracy increases with larger rank increment. Nevertheless, rank-1 increment provides us comparable or better performance than the comparing techniques as shown in Table~\ref{table:compare}.
% \vspace{-2mm}

% \begin{table}[h]
% \caption{ Test accuracy for different rank choices of the proposed ITL approach and multi-task baseline networks for P-MNIST, and R-MNIST. Initial rank is $r_{k,1}$ and rank increment/task is $r_{k,t}$.\ \\[-1.0ex]}\label{table:rank}
% \centering % centering table
% \small
% \begin{tabular}{l |c |c | c c  } % columns
% \hline\hline % inserting double-line
%  Method & $r_{k,1}$ & $r_{k,t}$ & \footnotesize{P-MNIST} &  \footnotesize{R-MNIST}  
% \\ 
% \hline % inserts single-line
% \textbf{Proposed ITL Setup-1}  & 1  & 1 & 74.23 & 81.57 \\
% \textbf{Proposed ITL Setup-2}  & 6  & 1 & 82.21 & 89.39 \\
% \textbf{Proposed ITL Setup-3}  & 11  & 1 & 85.61 & 91.09 \\
% \textbf{Proposed ITL Setup-4}  & 11  & 2 & 90.51 & 92.76 \\
% \textbf{Proposed ITL Setup-5}  & 11  & 4 & 93.84 & 94.12 \\
% \hline

% \end{tabular}

% \end{table}

\begin{table}[t]
% \vspace{-4mm}
\caption{ Test accuracy for different rank choices of the proposed ITL approach and multi-task baseline networks for P-MNIST and R-MNIST. Initial rank is $r_{k,1}$ and rank increment/task is $r_{k,t}$.\ \\[-1.0ex]}\label{table:rank}
\centering % centering table
\small
\begin{tabular}{l|cl|cl|cl|cl|cl}
\hline
Setup         & \multicolumn{2}{c|}{1}     & \multicolumn{2}{c|}{2}     & \multicolumn{2}{c|}{3}      & \multicolumn{2}{c|}{4}      & \multicolumn{2}{c}{5}      \\ \hline
    $(r_{k,1},r_{k,t})$          & \multicolumn{2}{c|}{(1,1)} & \multicolumn{2}{c|}{(6,1)} & \multicolumn{2}{c|}{(11,1)} & \multicolumn{2}{c|}{(11,2)} & \multicolumn{2}{c}{(11,4)} \\ \hline \hline
P-MNIST       & \multicolumn{2}{c|}{74.23} & \multicolumn{2}{c|}{82.21} & \multicolumn{2}{c|}{85.61}  & \multicolumn{2}{c|}{90.51}  & \multicolumn{2}{c}{93.84}  \\ \hline
R-MNIST       & \multicolumn{2}{c|}{81.57} & \multicolumn{2}{c|}{89.39} & \multicolumn{2}{c|}{91.09}  & \multicolumn{2}{c|}{92.76}  & \multicolumn{2}{c}{94.12}  \\ \hline \hline
\# parameters & \multicolumn{2}{c|}{0.09M} & \multicolumn{2}{c|}{0.1M}  & \multicolumn{2}{c|}{0.11M}  & \multicolumn{2}{c|}{0.14M}  & \multicolumn{2}{c}{0.2M}   \\ \hline
\end{tabular}
% \\[-3ex]
\end{table}

% \vspace{-4mm}

\subsection{Results with ResNet18} 
The proposed low-rank increments approach can be generalized to other type of networks and layers as well. For example, convolutional kernels have four-dimensional weight tensors as opposed to the two-dimensional weight matrices of fully connected layers. They are usually formulated as a tensor of output and input channel $(C_{out}, C_{in})$, and the two dimensions of the convolutional filters $(H, W)$. We reshape the convolutional weight tensors into matrices of size ${C_{out} \times  C_{in}HW}$ and perform similar low-rank updates per task as we described for the MLP in the main paper. We report the results for S-CIFAR-100 and S-miniImageNet datasets with Resnet18 architecture. For each convolutional layers, we reshaped and decomposed the convolution weight tensors into the same low-rank factors described in (\ref{eq:weight}) and performed low-rank updates per tasks. We report the results in Table~\ref{table:resnet}. For most of the comparing techniques, results from~\cite{chaudhry2020continual} are reported since we use the same architecture and dataset. For missing comparisons, we trained the models using same procedure as outlined in \cite{chaudhry2020continual}.

Instead of using a fixed value for rank at each layer as we did in the MLP setup, we used rank size that is proportional to the size of $C_{out, i}$ at $i^{th}$ convolutional layer because the weights for different layers of ResNet18 are different in size. We select initial rank = $0.1\,C_{out, i}$ for the first task and incremental rank = $0.02\,C_{out, i}$ for the subsequent incremental tasks.

The results in Table~\ref{table:resnet} show that the performance of every method improves with the convolutional ResNet18 structure over the 3-layer MLP. Nevertheless, our method outperforms the comparing approaches for both datasets. Adam-NSCL \cite{wang2021training} gets better results on CIFAR100, but it requires 11.21M parameters (compared to 1.33M parameters required by our method). 

% \Salnote{Do we have parameter count for other methods with Resnet18?}

\begin{table*}[h]
\centering
% \vspace{-5mm}
\caption{Comparison of test accuracy and forgetting for split CIFAR-100 and split miniImageNet datasets using ResNet18 architecture. 
 }\label{table:resnet}
\small
\begin{tabular}{l|cc|cc}
\hline
\hline
%\\ [-1ex]
Method & \multicolumn{2}{c}{S-CIFAR-100}
& \multicolumn{2}{c}{S-miniImageNet}
\\
& Accuracy& Forgetting
& Accuracy& Forgetting\\
\hline
EWC \cite{kirkpatrick2017overcoming} &43.2 ($\pm$2.77) &26 ($\pm$2)&34.8 ($\pm$2.34)& 24 ($\pm$4)\\
ICARL \cite{rebuffi2017icarl} &46.4 ($\pm$1.21) & 16 ($\pm$1) & 44.2 & 24.64\\
AGEM \cite{chaudhry2018efficient} & 60.34 ($\pm$2.05) & 11.0 ($\pm$2.88) &42.3 ($\pm$1.42) & 17 ($\pm$1))\\
ER-Ring \cite{chaudhry2019tiny} &59.6 ($\pm$11.9) &0.14 ($\pm$1) &49.8 ($\pm$2.92) &12 ($\pm$1)\\
Ortho sub \cite{chaudhry2020continual}  & 63.42 ($\pm$1.82) & 8.37 ($\pm$0.71) &51.4 ($\pm$1.44) &10 ($\pm$1)\\
% DER & 67.16 & ?? & 57.81 & ?? \\
Adam-NSCL \cite{wang2021training} & \textbf{74.31} & 9.47 & 57.92 & 13.42 \\ 
IBP-WF \cite{mehta2021continual} & 68.25 & 0 & 55.84 & 0 \\ 
Ours & 68.46 ($\pm$2.52)  & \textbf{0}
& \textbf{59.26} ($\pm$1.15)  & \textbf{0}\\      
\hline
Parallel full-rank & 92.7 &0 & 94.5 &0
\\\hline
Multitask learning & 70.2 & 0 &65.1&0\\\hline
\end{tabular}

\end{table*}

\noindent \textbf{Effect of updating last few layers.}  
We performed an experiment on S-CIFAR-100 where we factorize last $L$ layers of the ResNet18 architecture keeping the rest of the network fixed at trained weights on Task 1. Updating last $L=\{1,2,3,4,5\}$ layers provide average accuracy of  $\{34.38,  34.99, 53.41, 57.08, 65.03\}$, respectively. This result suggests that updating last few layers may suffice since the initial layers merely work as a feature extractor.